\documentclass[11pt,letterpaper]{article}
\usepackage[margin=1in]{geometry}
\usepackage[utf8]{inputenc}
\usepackage[T1]{fontenc}
\usepackage{lmodern}
\usepackage{amsmath,amsfonts}
\usepackage{graphicx}
\usepackage{booktabs,tabularx,array}
\usepackage{microtype}
\usepackage{xcolor}
\usepackage{xurl}
\usepackage{natbib}
\usepackage[hidelinks]{hyperref}
\usepackage{placeins}
\usepackage{needspace}
\usepackage{fancyvrb}
\usepackage[font=small,labelfont=bf]{caption}
\setcitestyle{authoryear,square,semicolon}
\hypersetup{
  pdftitle={Persistent Memory and Iterative Improvement in an LLM Agent for Chinese-Character Drawing},
  pdfauthor={Peilin Wu and Yi-Zhuang You},
  pdfsubject={Standalone research preprint on persistent memory in LLM agents}
}
\title{Persistent Memory and Iterative Improvement\\in an LLM Agent for Chinese-Character Drawing}
\author{%
  Peilin Wu\thanks{First author.}\\
  Palo Alto High School\\
  \texttt{peilinw2008@gmail.com}
  \and
  Yi-Zhuang You\thanks{Senior author.}\\
  University of California San Diego\\
  \texttt{yzyou@ucsd.edu}
}
\date{October 2026}
\begin{document}
\maketitle

\begin{abstract}
Does persistent memory accumulated from previous attempts help an LLM agent solve new problems, and does memory design affect that benefit? We study programmatic Chinese-character drawing across five agent conditions attempting the same 668 items. Accumulated memory produces modest aggregate gains despite substantial growth in the memory stores: free-form notes improve first-attempt success by 4.6 percentage points, while a design combining free-form memory with reusable code improves it by 2.8 points. The two designs do not differ significantly overall, but the code bank shows a suggestive advantage on later, harder items. A separate post-hoc comparison finds a much larger improvement when the code-bank agent receives privileged, target-specific structural supervision. The agent can therefore exploit highly informative structural guidance, although the privileged per-item information does not establish what could have been learned from experience alone. These results suggest that memory design may affect where accumulated experience becomes useful, even when its average benefit is modest.
\end{abstract}

\section{Introduction}
\subsection{Motivation}
A recurring claim in AI research is that persistent, updatable memory should let agents improve open-endedly on tasks where a single forward pass is not enough. The claim underpins much of the current interest in agent-driven science \citep{boiko2023}: give the system a hard problem, let it attempt, observe failure, write to memory, and converge. For the claim to be actionable, two questions matter. Does persistent memory accumulated from previous attempts improve performance on new items relative to the same agent without memory? If so, does the design of that memory change how useful past experience becomes?

Direct observation on frontier scientific tasks is difficult: ground truth is scarce and per-item cost is high. We instead need a proxy task that is (a) out-of-distribution enough that recall is not the story, (b) compositional enough that memory has room to help, and (c) quality-gradable, so that ``correct'' and ``correct and well-formed'' can be distinguished---the two-tier judgment real scientific artifacts receive but benchmarks rarely measure: a proof can be valid but inelegant, a synthesis correct but impractical.

\subsection{Chinese-character drawing as a proxy}
Writing Python programs that render Chinese characters as $300\times300$ images satisfies all three criteria. The task requires generating executable rendering code for each target character at pixel fidelity, rather than simply identifying the glyph; recent benchmarks find general-purpose models score near zero on text-to-glyph generation \citep{vecglypher}, making this a challenging compositional problem. Characters decompose into radicals and strokes that recur across the curriculum, so effective memory should crystallize component knowledge and reuse it. And a human judge can grade the same rendering on correctness and on calligraphic quality, giving the two-tier signal we need.

To answer these questions, we contribute a controlled five-condition experiment (668 items per condition; 3,798 blind curriculum judgments\footnote{$668\times5$ first attempts (3,340) plus 458 retry attempts. The original released log totals 3,948 rows, including 150 pilot-phase and re-judgment rows excluded from analysis.}) with matched contrasts separating memory-format comparisons from structural-supervision access. Accumulated memory yields modest aggregate gains, while the code bank shows a suggestive later/harder-item advantage. The post-hoc comparison shows that the agent can exploit supplied structural guidance, without establishing that equivalent guidance could have been learned from experience alone. We release attempts, judgments, memory snapshots, and protocol changes, and analyze retry behavior alongside first-attempt outcomes.

\section{Method}
\subsection{Task, curriculum, and batches}\label{sec:task}
The curriculum contains 668 items in two phases: 135 radicals (component sub-characters) followed by 533 full characters, ordered by ascending stroke count---from single-stroke radicals to characters of ten or more strokes. Ordering by stroke count means difficulty rises with curriculum position; batch index and difficulty are therefore deliberately coupled (we return to this in \S\ref{sec:limitations}).

Memory-equipped groups use two agent roles, both running Claude Opus 4.7: a drawer, which produces images, and a curator, which reads verdicts and maintains memory. Drawers never write memory and curators never draw; the memoryless control uses only a drawer. For each item, the drawer receives the character label, a ground-truth reference image, and, where applicable, read access to its group's memory (described next). It writes a Python program using the PIL graphics library, executes it to produce a $300\times300$ image, and may revise once after visually comparing its output to the reference. Throughout, an \emph{attempt} is one drawer dispatch yielding one judged image---first-attempt numbers therefore include that single internal revision---and a \emph{retry} is a fresh attempt of a previously failed item in a later batch (\S\ref{sec:architecture}).

Items are processed in batches of 50 (the opening batch of 18 radicals excepted). Batching serves two purposes. Scientifically, memory is frozen within a batch: all 50 items are attempted under the identical memory state, which makes per-batch numbers interpretable. Practically, it lets the human judge grade in sittings and the memory system update on a feedback cycle, rather than requiring a judge--update round trip after every item.

After each batch, a human judge grades every attempt (protocol in \S\ref{sec:judgment}). For the first eight batches the rubric was binary (Pass/Fail); at batch B9 we refined it into four tiers---Perfect (calligraphically clean, print quality), Pass (recognizable and correct), Close (nearly correct), Fail---once it became clear that a correctness-only signal could not distinguish crude success from mastery.\footnote{In the released data files these tiers appear under their original run-time labels A, PASS, C, FAIL.} After judging, each memory-equipped group's curator updates memory and queues failed items for retry (\S\ref{sec:architecture}).

\subsection{The five conditions}\label{sec:conditions}
All five groups attempt the same 668 items using the common drawer harness. Their memory designs, rendering representations, and access to structural supervision are described below.

\textbf{G1---no memory (control).} The drawer receives only the label and reference image. There is no curator; nothing persists between items. G1 measures what the base model can do cold, item after item.

\textbf{G2---free-form notes.} Memory is a directory of markdown files the curator writes and organizes however it likes: error patterns, per-radical advice, pre-drawing checklists. The drawer reads these notes before attempting. This is the ``lessons-learned notebook'' format, in the spirit of reflection-style agent memory.

\textbf{G3---free-form memory + code bank.} The original three-bank architecture includes a free-form persistent Sandbox, a Principle Bank, and a Success Bank of callable Python functions promoted from human-judged successful attempts. The drawer may import and call these functions instead of drawing from scratch. We call a promoted, reusable drawing function a \emph{primitive}; it may render a stroke, radical, or whole character.

\textbf{G4---free-form memory + grid bank + supervision.} G4 also includes free-form persistent memory from the outset. Its structured bank stores stroke and radical knowledge in a mizige grid-coordinate language, based on the Chinese handwriting practice grid with horizontal, vertical, and diagonal guides. The drawer uses a width-modulated primitive, \texttt{fat\_line(w1,w2)}, that can taper strokes between endpoints. G4 additionally receives the per-attempt structural supervision described in \S\ref{sec:supervision}; this injection was discovered during the run (\S\ref{sec:error}).

\textbf{G5---free-form memory + code bank + supervision.} The same memory design as G3, plus the same per-attempt structural supervision as G4. G5 was added late in the run to supply the missing matched comparison (\S\ref{sec:error}); it was run through the full curriculum from the beginning.

The primary comparisons are G1 versus G2/G3 for the benefit of accumulated memory, and G2 versus G3 for memory format without structural supervision. G3 versus G5 provides the post-hoc structural-supervision diagnostic under a fixed code-bank format. G4 versus G5 compares memory-and-rendering bundles rather than memory format alone.

\subsection{Memory architecture and retries}\label{sec:architecture}
Drawers never write to memory; curators never draw. Alongside its memory files, each memory-equipped group maintains an auto-generated index of every successful attempt with a pointer to the rendered image, so drawers can consult prior successes visually. For the first 150 items memory organization was fixed; from midway through batch B3 onward, curators could freely restructure their own files---create, split, retire---subject only to the group's core format constraint, and each group's reorganizations were logged. From item 350 onward, the obligation to reuse bank functions was lifted: imports remained fully available, but drawers could equally render fresh inline code, a change made after we observed drawers deferring to imperfect bank entries. From item 500, a deviation channel let a drawer skip a bank function with a note to the curator, and passing deviations could be promoted as variants.

Failed items can be retried under rules identical for all memory-equipped groups: every 25 items the curator scans its error list and self-selects retries it can justify (an upcoming prerequisite, or a specific new insight addressing the failure); a retried item must wait 50 further curriculum items before another retry; and items still unsolved when the curriculum ends are frozen. There is no fixed per-item retry cap---retry volume is the curator's own judgment. \S\ref{sec:retries} analyzes this subsystem.

\subsection{Target-specific structural supervision}\label{sec:supervision}
The supervision is extracted from the same MakeMeAHanzi stroke medians \citep{mmh} used to render the reference images shown to all groups. At dispatch time, G4 and G5 receive a ``structural expectations'' block specifying the target's stroke count, per-stroke head and tail coordinates at roughly pixel precision, and joint types: welded, small-gap, or tangent (Appendix~\ref{app:memory}). This is privileged, target-specific information rather than accumulated experience. It supplies neither intermediate curve shapes nor stroke widths, which the drawer must determine. No group may read the underlying database directly.

\subsection{Judgment}\label{sec:judgment}
\begin{figure}[tbp]
\centering
\includegraphics[width=\linewidth]{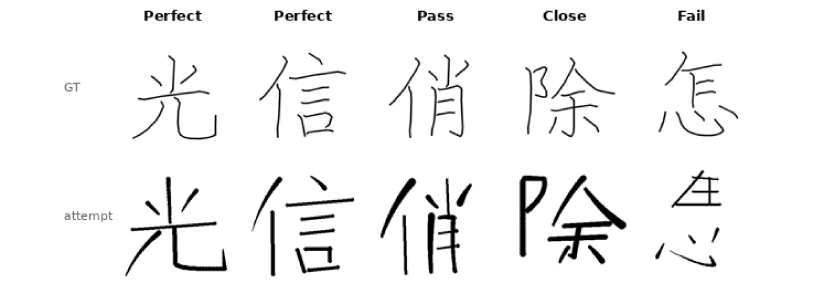}
\caption{The verdict tiers, real attempts (bottom row) with their references (top row), drawn from the four-tier window (B9--B13). Perfect (two examples): correct and calligraphically modulated. Pass: correct but crude. Close: nearly right (here the left radical collapses into a blob). Fail: components present but misassembled.}
\label{fig:tiers}
\end{figure}

One human (the first author) graded every attempt blind: a shuffling tool presents only the character label, the reference image, and a single anonymized attempt. Perfect is stringent---correct strokes rendered with uniform-width crudeness earn Pass, not Perfect. \emph{Success} throughout the paper means Perfect or Pass. Figure~\ref{fig:tiers} shows real examples of each tier, drawn from the four-tier window.

A second rater independently graded a frozen random sample of 150 first attempts from B9--B13, 30 per condition, using the same rubric and blind to condition, original verdict, and study hypotheses. Success/Failure agreement was 92.7\% (Cohen's $\kappa=.827$, 95\% CI $[.728,.925]$); Perfect/non-Perfect agreement was 96.0\% ($\kappa=.730$). Disagreements were not adjudicated, and the original experimental labels remain unchanged.

\FloatBarrier
\section{Results}
\subsection{Trajectories}\label{sec:trajectories}
\begin{figure}[tbp]
\centering
\includegraphics[width=\linewidth]{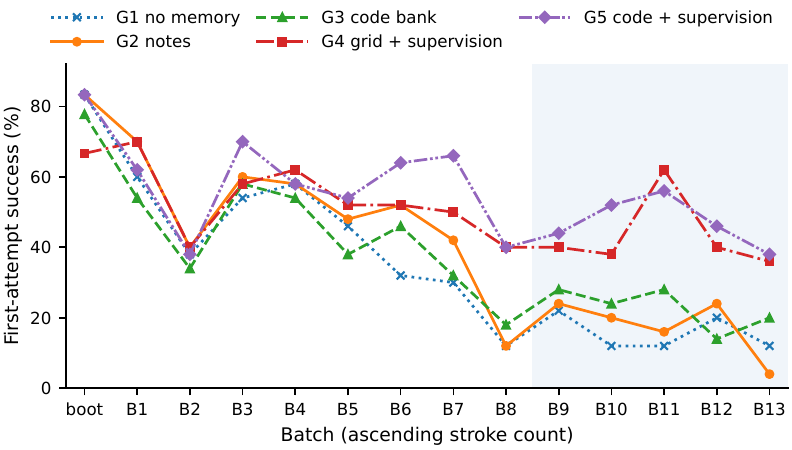}
\caption{First-attempt success by batch; ``boot'' contains 18 items. Distinct markers and line patterns identify the five conditions. Shading marks B9--B13, the common four-tier evaluation window.}
\label{fig:trajectories}
\end{figure}

Figure~\ref{fig:trajectories} shows per-batch success (Perfect+Pass). All groups start at 60--83\% on radicals; over the six deepest batches (B8--B13) the three groups without structural supervision run at 4--28\% while the two supervised groups sustain 36--62\%.

\subsection{Cumulative outcomes}\label{sec:outcomes}
Table~\ref{tab:outcomes} reports first-attempt success across all 668 items and Perfect verdicts within the common B9--B13 window: items 419--668, comprising 250 items per condition. The complete four-tier distribution is reported only for this common window (Appendix~\ref{app:fourtier}).

\begin{table}[tbp]
\caption{First-attempt outcomes. Five of G1's six Perfect outputs were rendered with a system font rather than drawn; the original image verdicts are retained, with provenance documented in Appendix~\ref{app:shortcut}.}
\label{tab:outcomes}
\centering
\begin{tabular}{@{}lrrrr@{}}
\toprule
 & \multicolumn{2}{c}{Success, all 668} & \multicolumn{2}{c}{Perfect, B9--B13 ($n=250$)}\\
\cmidrule(lr){2-3}\cmidrule(l){4-5}
Condition & $n$ & Rate & $n$ & Rate\\
\midrule
G1 no memory & 219 & 32.8\% & 6 & 2.4\%\\
G2 notes & 250 & 37.4\% & 5 & 2.0\%\\
G3 code bank & 238 & 35.6\% & 1 & 0.4\%\\
G4 grid + supervision & 332 & 49.7\% & 51 & 20.4\%\\
G5 code bank + supervision & 359 & 53.7\% & 38 & 15.2\%\\
\bottomrule
\end{tabular}
\end{table}

\subsection{Accumulated memory and memory design}\label{sec:memoryresults}
Relative to G1, free-form notes increase first-attempt success by 4.6 percentage points (discordant pairs 80 vs.\ 49; exact McNemar $p=.008$), while G3's increase is 2.8 points ($p=.129$). The direct comparison between G2 and G3 is not significant: 37.4\% versus 35.6\%, with 69 versus 57 discordant successes ($p=.327$). These are modest aggregate benefits despite substantial growth in the memory stores (Table~\ref{tab:checkpoints}).

Within the common quality window, G2 and G3 receive five and one Perfect verdicts, respectively. Four items are Perfect only for G2 and none only for G3 (exact McNemar $p=.125$), providing insufficient evidence for a quality advantage of either design.

\begin{table}[tbp]
\caption{Observed memory checkpoints. G2 counts curator-authored operational memory, excluding automatically generated indexes and process logs; G3/G4 count bank primitives. These are architecture-specific units, not comparable information quantities. G4's final decrease reflects pruning.}
\label{tab:checkpoints}
\centering
\begin{tabular}{@{}lrrr@{}}
\toprule
Curriculum position & G2 lines & G3 primitives & G4 primitives\\
\midrule
0 & 0 & 0 & 0\\
$\approx150$ & 1,011 & 67 & 78\\
$\approx400$ & 3,707 & 184 & 140\\
668 & 5,111 & 251 & 130\\
\bottomrule
\end{tabular}
\end{table}

\subsection{Relative benefit across the curriculum}\label{sec:relative}
\begin{figure}[tbp]
\centering
\includegraphics[width=\linewidth]{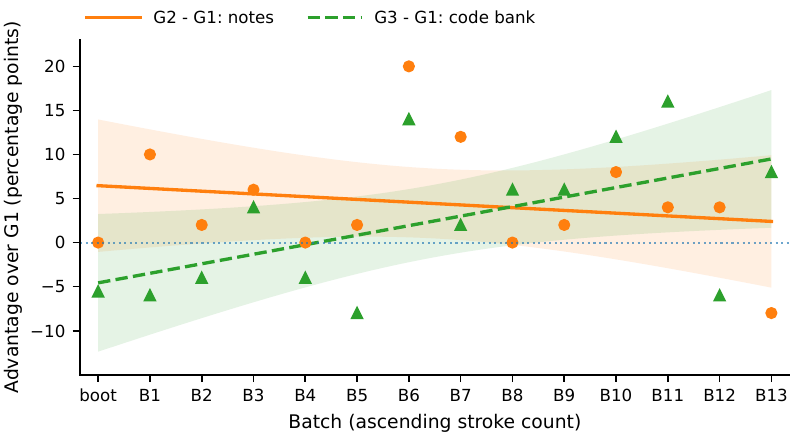}
\caption{Batch-matched advantage over the memoryless control. Points show observed success-rate differences; lines show linear fits with pointwise 95\% confidence bands. Circles denote G2$-$G1 and triangles G3$-$G1. Bootstrap contains 18 items; subsequent batches contain 50.}
\label{fig:advantage}
\end{figure}

Figure~\ref{fig:advantage} plots each memory condition's success rate minus G1's on the same batch items. Over B8--B13, G3 succeeds on 22.0\% of 300 items, compared with 16.7\% for G2 and 15.0\% for G1. Paired comparisons give $p=.040$ for G3 versus G2 and $p=.0075$ for G3 versus G1; G2 versus G1 is not significant ($p=.542$).

An exploratory, post-hoc linear fit across the 14 batches gives a slope of $+1.08$ percentage points per batch for G3$-$G1 ($p=.040$), compared with $-0.31$ for G2$-$G1 ($p=.502$). This is suggestive of a later/harder-item advantage for the code-bank design, subject to the curriculum coupling described in \S\ref{sec:limitations}.

\subsection{Target-specific structural supervision}\label{sec:supervisionresults}
On B9--B13, G5 exceeds G3 in first-attempt success: 47.2\% versus 22.8\%, a gain of 24.4 percentage points (discordant pairs 74 vs.\ 13; exact McNemar $p=1.6\times10^{-11}$). Perfect verdicts increase from 1/250 to 38/250 ($p=1.5\times10^{-10}$).

The complementary bundle comparison, G4 versus G5, gives a numerically higher Perfect rate for G4 (20.4\% vs.\ 15.2\%, $p=.072$) but lower success (43.2\% vs.\ 47.2\%, $p=.237$). Neither difference is significant.

Across the attempt scripts covered by the code audit, bank imports appear in 46.1\% of G3 attempts and 91.0\% of G5 attempts. These rates describe explicit imports, not all uses of stored memory.

\subsection{Retries and freezing}\label{sec:retries}
Table~\ref{tab:retries} partitions initially failed items into never retried, retried but unresolved, and recovered. Under the same retry instructions, G2 and G3 retry 10.8\% and 13.0\% of their failures (Fisher's exact $p=.341$). G2 recovers 12/45 retried items (26.7\%), compared with 6/56 for G3 (10.7\%; $p=.065$). These comparisons do not establish a memory-design effect on retry appetite or recovery.

G5 retries a larger share of its failures than G3 (34.6\% vs.\ 13.0\%, $p=4.2\times10^{-12}$) and recovers a larger share of selected items (25.2\% vs.\ 10.7\%, $p=.039$). G4 retries fewer failures than G5 (25.9\% vs.\ 34.6\%, $p=.016$), with a numerically higher recovery rate (37.9\% vs.\ 25.2\%, $p=.063$). These contrasts are reported separately rather than pooled as a format effect. Recovery rates are conditional on each curator's selected items.

Including retries, final success is G5 57.8\%, G4 54.6\%, G2 39.2\%, G3 36.5\%, and G1 32.8\%. Retries add 0.9--4.9 percentage points for the memory-equipped groups without changing their ordering.

\begin{table}[tbp]
\caption{Retry outcomes. Retried is a share of failed items; Recovered is a share of retried items; Perfect counts failed items recovered to the top tier; Frozen and Never retried are shares of failed items. The control (449 failures) has no curator and hence no retries. G4 additionally re-attempted five already-passing items hunting Perfect upgrades (two succeeded); these are excluded from the failure partition. Diagnostic reruns under altered prompts are excluded from this accounting.}
\label{tab:retries}
\centering
\small
\begin{tabular}{@{}lrrrrrr@{}}
\toprule
Group & Failed & Retried & Recovered & Perfect & Frozen & Never retried\\
\midrule
G2 notes & 418 & 45 (10.8\%) & 12 (26.7\%) & 0 & 33 (7.9\%) & 89.2\%\\
G3 code bank & 430 & 56 (13.0\%) & 6 (10.7\%) & 0 & 50 (11.6\%) & 87.0\%\\
G4 grid + sup. & 336 & 87 (25.9\%) & 33 (37.9\%) & 4 & 54 (16.1\%) & 74.1\%\\
G5 code + sup. & 309 & 107 (34.6\%) & 27 (25.2\%) & 3 & 80 (25.9\%) & 65.4\%\\
\bottomrule
\end{tabular}
\end{table}

\begin{figure}[tbp]
\centering
\includegraphics[width=\linewidth]{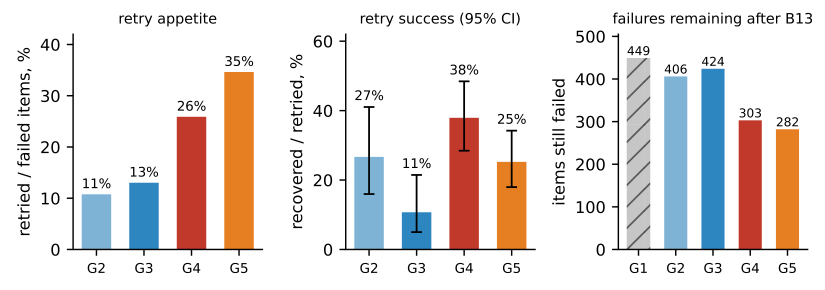}
\caption{Retry behavior. Left: proportion of initially failed items selected for retry. Center: proportion of selected items recovered, with 95\% confidence intervals. Right: items remaining unsuccessful after the curriculum.}
\label{fig:retries}
\end{figure}

\FloatBarrier
\section{Discussion}
\subsection{What this says about accumulated memory}\label{sec:discussion}
The later code-bank advantage changes how the aggregate comparison should be read. It motivates evaluating memory designs across compositional demands rather than ranking them by a single overall score. A design containing executable components may become useful in different parts of a curriculum from one storing only prose, even when their average performance is similar.

G5 provides a separate diagnostic: the agent can exploit an explicit structural decomposition when it is supplied. The memory-only conditions did not develop representations with comparable observed utility. Because the supplied decomposition contains privileged information about each target, however, this comparison does not establish that equivalent guidance could have been learned from experience alone.

The potentially transferable finding is the difference between average memory benefit and benefit on demanding compositional items. Tasks involving recurring components offer a reason to test this pattern elsewhere; the present experiment does not establish that it holds across niche tasks generally.

\subsection{The implementation error behind G5}\label{sec:error}
Late in the run (after batch B11) we discovered the source of an asymmetry: G4's curator had misread its setup instructions during configuration, so the dispatch pipeline had been prepending the structural block of \S\ref{sec:supervision}---the structural supervision---to G4's briefs and to no other group's. This implementation error would have left the memory-format comparison uninterpretable. Rather than alter G4 mid-run and split its trajectory, we left it unchanged and added G5: G3's exact memory format plus the same injection, run through the full curriculum (\S\ref{sec:supervisionresults}).

\subsection{Limitations}\label{sec:limitations}
All agents ran on a single model, Claude Opus 4.7 (including G5's catch-up run, occurring one week after the other groups'), on a single task whose quality dimension---calligraphic modulation---is task-specific; we do not claim generality across models or domains. Curriculum position and difficulty are coupled by design (\S\ref{sec:task}), so trajectories cannot separate ``later in training'' from ``harder items.'' Finally, the protocol evolved during the run; every change is catalogued in Appendix~\ref{app:ledger}.

\subsection{Related work}
A prominent line of work equips LLM agents with accumulating memory and reports iterative gains: Voyager grows a library of executable skills in an open-ended environment \citep{voyager}, Reflexion stores verbal self-reflections that inform later trials \citep{reflexion}, ExpeL distills reusable insights from experience \citep{expel}, Agent Workflow Memory induces reusable workflows \citep{awm}, and SkillWeaver lets web agents discover and hone skills \citep{skillweaver}. A parallel line builds the memory substrate itself---MemGPT's paged hierarchy \citep{memgpt}, generative agents' retrieval-scored event streams \citep{generative}, structured long-term stores such as HippoRAG \citep{hipporag}---and recent surveys map the space \citep{surveyzhang,surveyluo}. Our study compares free-form and code-augmented persistent-memory designs on a shared quality-graded curriculum, with target-specific structural supervision examined separately. On the task side, benchmarks find general-purpose models weak at low-level vector graphics \citep{vgbench} and near zero at text-to-glyph generation \citep{vecglypher}; where prior work trains specialists for glyph drawing, we ask whether a general agent can climb the same hill through memory and supervision.

\section{Conclusion}\label{sec:conclusion}
Persistent accumulated memory produced modest aggregate gains despite substantial growth in the memory stores: 4.6 percentage points for free-form notes and 2.8 points for the code-augmented design. Although the designs did not differ significantly overall, the code bank showed a suggestive advantage on later, harder items. Memory design may therefore influence where experience helps even when its average benefit is small.

In a separate post-hoc diagnostic, target-specific structural supervision produced a 24.4-point gain on the common quality window. The agent could exploit useful structure when supplied; whether memory systems can learn to construct comparably useful structure from experience remains open.

\section*{Data availability}
All judgments (\texttt{item\_grades.csv}), attempts, memory snapshots, the protocol-change ledger, per-group retry logs recording both attempted and declined retries with the curator's stated reason, and analysis code are released at \url{https://github.com/Clearsightpei/iterative-agent-memory}. We do not redistribute MakeMeAHanzi's data; reproducers obtain it upstream \citep{mmh} at commit \texttt{618dbab8} (our copy verified byte-identical); its Arphic-lineage license texts accompany our release, together with our derived per-character joint tables and tooling.
\label{maintext:end}
\FloatBarrier
\clearpage

\section*{Acknowledgments}
P.W. designed and implemented the experiment, performed all 3,798 curriculum judgments, ran the analysis, and wrote the paper. Y.-Z.Y. proposed the research direction and provided guidance and feedback throughout.

\bibliographystyle{unsrtnat}
\bibliography{references}

\appendix
\section{Protocol-change ledger}\label{app:ledger}
Every mid-run change, in curriculum order. Positions are item indices (668 total).

\begin{table}[h]
\centering
\begin{tabularx}{\linewidth}{@{}p{0.65in}>{\raggedright\arraybackslash}X>{\raggedright\arraybackslash}X@{}}
\toprule
Position & Change & Reason\\
\midrule
150 & Curators may freely restructure their own memory files within format constraints & Fixed organization was bottlenecking G2/G3\\
270 & Retry cool-down and terminal freeze introduced & Immediate re-attempts were wasting budget\\
350 & Obligation to reuse bank functions lifted; imports still available, fresh inline rendering equally allowed & Drawers deferred to imperfect bank entries\\
419 & Four-tier rubric (Perfect/Close added) from batch B9 & Binary rubric could not distinguish crude success from mastery\\
500 & Bank-deviation channel with curator note & Let drawers bypass bad primitives without silent divergence\\
post-B11 & G5 added (G3 format + supervision); G4 left unchanged & G4's curator misread its setup instructions; injection found reaching G4 only (\S\ref{sec:error})\\
\bottomrule
\end{tabularx}
\end{table}

\section{Memory artifacts (verbatim excerpts)}\label{app:memory}
G2 note (\texttt{sibling\_signature\_checklist.md}):
\begin{Verbatim}[fontsize=\small]
# Sibling Signature Checklist (pos 277)
Created in response to B4 retry-rate drop (31% -> 21%).
Every retry FAIL in B4 traces to either (a) a signature-bit
 override (draw the wrong sibling) or (b) a terminal-flick
 omission. Consult BEFORE drawing any item listed below...
\end{Verbatim}
G3 bank entry (\texttt{ba.py}):
\begin{Verbatim}[fontsize=\small]
# ba.py -- "eight" radical, 2 strokes (pie + na, splayed
# with a V-notch gap). Bootstrap batch -- human PASSed.
from pie import draw_pie
from na import draw_na
def draw_ba(t, ox=0, oy=0, scale=1.0): ...
\end{Verbatim}
Injected structural block (as received by G4/G5 drawers):
\begin{Verbatim}[fontsize=\small]
Character: <target>
Expected stroke count: 4 (generated.py must produce
 exactly this many strokes)
Stroke 1: head (72,41) -> tail (105,214) joint J1->2: welded
Stroke 2: head (105,214) -> tail (228,199) joint J2->3: gap
...
\end{Verbatim}
Operational-memory counts include curator-authored files available to support drawing, excluding the automatically generated \texttt{pass\_index.md} and process records in \texttt{evolution.md} and \texttt{scans/}. At the final checkpoint, G2 contains 5,111 lines: 2,556 in \texttt{errata.md} and 2,555 across the remaining seven files.

\section{Shortcut items}\label{app:shortcut}
A code audit found that five of G1's six Perfect outputs, together with two B8 passes, were rendered using a system Chinese font rather than composed from strokes. No audited memory-equipped attempt used this shortcut. The original blind image verdicts are retained. The pattern is consistent with procedural memory constraining strategy selection, but remains a descriptive observation rather than an isolated causal effect \citep{amodei,krakovna,skalse}.

\begin{figure}[h]
\centering
\includegraphics[width=.70\linewidth]{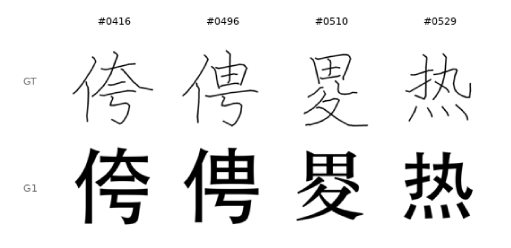}
\caption{The four items where the memoryless control alone exceeded Close. Top: reference. Bottom: the control's render---typeset glyphs via system font, not drawn strokes.}
\label{fig:shortcut}
\end{figure}

Indices are character-phase item ordinals; absolute curriculum position is the ordinal plus 135 (item 416 sits at position 551, batch B11). Control-group font-rendered items: two B8 passes (indices 258, 274) and five Perfects at indices 416, 482, 496, 510, 529; the genuinely drawn control Perfect is index 354 (B10). Four of the five font Perfects appear in Figure~\ref{fig:shortcut}; index 482 does not, because that figure's criterion is items where no other group exceeded Close, and on 482 both supervised groups did (G4 Pass, G5 Perfect).

Removing the seven font-rendered control items gives $212/661=32.1\%$ success, compared with the reported $219/668=32.8\%$. The main results retain the original blind image verdicts.

\section{Common-window four-tier outcomes}\label{app:fourtier}
\begin{table}[h]
\caption{Four-tier first-attempt outcomes on the same 250 items from B9--B13 for every condition.}
\centering
\begin{tabular}{@{}lrrrr@{}}
\toprule
Condition & Perfect & Pass & Close & Fail\\
\midrule
G1 & 6 & 33 & 25 & 186\\
G2 & 5 & 39 & 34 & 172\\
G3 & 1 & 56 & 34 & 159\\
G4 & 51 & 57 & 41 & 101\\
G5 & 38 & 80 & 33 & 99\\
\bottomrule
\end{tabular}
\end{table}

\section{Additional judgment-reliability results}\label{app:reliability}
Four-tier inter-rater agreement was $\kappa=.554$ unweighted, $.721$ with linear weights, and $.846$ with quadratic weights. Of 43 disagreements, 41 were between adjacent tiers and 26 were between Fail and Close, leaving the binary outcome unchanged in those cases. The second rater confirmed all nine original Perfect labels in the sample and assigned six additional Perfect labels.

\FloatBarrier
\end{document}